\documentclass[]{fairmeta}
\usepackage{amsmath,amssymb,amsthm}
\usepackage[ruled,vlined]{algorithm2e}
\usepackage{enumitem}

\newtheorem{lemma}{Lemma}
\newtheorem{corollary}{Corollary}

\SetAlgoSkip{}
\title{Exact Collision Gradients for Ellipsoids in Differentiable Simulation}

\author[1]{Shamak Dutta}
\author[1]{Biswadip Dey}

\affiliation[1]{Meta}

\abstract{Differentiable simulation enables gradient-based optimization in learning and control problems involving collisions, but its gradients are least reliable at the moment of impact, when the velocities jump. A fixed-step simulator detects a collision only after the bodies overlap, and some differentiable simulators correct for this with a linear estimate of the time of impact. In this paper, we study impacts between ellipsoids. We show that the gradient error of this estimate grows with the obliqueness of the collision, the curvature at contact, and the time step, and that ignoring spin adds a further error. Subsequently, we find the time of impact with continuous collision detection on the Perram--Wertheim contact function and differentiate through the touching condition rather than the search. This gives exact gradients at any step size. We demonstrate our method on an optimal control problem and a system identification problem. In both, exact gradients recover the optimal control and the true parameters, while the linear estimate does not, even when it accounts for spin.}

\date{\today}
\correspondence{Shamak Dutta at \email{shamak@meta.com}}
\metadata[Keywords]{Differentiable simulation, rigid-body impacts, time of impact, contact model}

\begin{document}

\maketitle

\section{Introduction}

Differentiable physics simulators compute gradients of a simulated outcome with respect to initial states, controls and physical parameters, which allows them to be used inside gradient-based optimization in robotics and control. They have been successfully applied to system identification, trajectory optimization, policy optimization and morphology optimization \citep[Section~IV]{newbury2024review}. One of the main challenges is contact. It is hard to model, and simulators do not always reproduce real impacts accurately \citep{acosta2022validating}. It is also hard to differentiate. A rigid impact changes the velocities instantly, so the gradient of the simulated outcome depends on how the simulator resolves the impact within a time step, and common choices can give wrong or unreliable gradients \citep{stewart2010optimal,zhong2022differentiable,suh2022differentiable}.

A fixed-step simulator detects a collision only after the bodies overlap. Some differentiable simulators correct for this by rewinding the bodies to a one-step linear estimate of the time of impact, as in DiffTaichi \citep{hu2020difftaichi} and its extension TOI-Velocity \citep{zhong2023improving}. This estimate is exact only when the gap between the bodies closes at a constant rate over the step, as in a head-on collision between spheres. However, many simulated bodies are not spheres, and when they collide obliquely or spin, the gap closes at a varying rate. The estimated impact time, contact point and normal are then wrong, and so are the gradients computed through them.

Ellipsoids are a standard non-spherical particle shape in discrete element models \citep{lin1997three}, and 3D Gaussian splatting represents scenes with anisotropic Gaussians whose level sets are ellipsoids \citep{kerbl2023gaussian}. Recent work couples Gaussian splatting with the material point method (MPM) \citep{xie2024physgaussian} and differentiates through the simulation to estimate physical parameters from video \citep{cai2024gic,rho2025projo4d,rho2026monophysics}. MPM suits deformable materials, but it handles contact on a grid and does not compute when or where two rigid bodies touch. For rigid objects, \citet{wang2026contactgaussian} use the Gaussians as collision geometry but restrict them to spheres to keep collision detection differentiable. Exact, differentiable contact between ellipsoids would remove this restriction and give a basis for coupling rigid-body simulation with Gaussian rendering.

In this paper, we study impacts between ellipsoids. Our contributions are as follows.
\begin{itemize}[nosep]
\item We show that the gradient error of the linear time-of-impact estimate grows with the obliqueness of the collision, the curvature at contact and the time step (Lemma~1). Ignoring spin, as TOI-Velocity does, adds an additional error.
\item We compute the exact time of impact between ellipsoids inside a fixed-step simulator using Perram--Wertheim contact function \citep{perram1985statistical} and differentiate through the touching condition to yield exact gradients.
\item We compare our exact gradients with those of TOI-Velocity and its extension to spin in optimal control and system identification experiments. Our gradients recover the optimal control and the true parameters, while neither baseline does.
\end{itemize}
\section{Problem Formulation}\label{sec:problem}

\subsection{Motivating Example} \label{subsec:motivation}
\begin{figure}[b!]
  \centering
  \includegraphics[width=\linewidth]{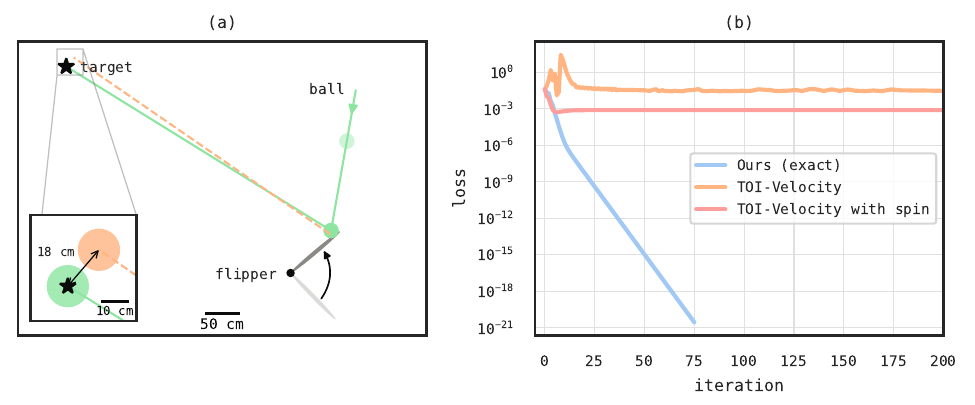}
  \caption{\small{(a) The pinball optimal control task as a motivating problem. The flipper rests at a fixed angle (faded), then swings to strike the ball, which travels to the target. Dashed: the ball's path under the final control found with TOI-Velocity's gradients, which misses the target by 18\,cm (inset). (b) True loss during gradient descent from the same initial control. With our gradients the loss decreases to about $10^{-21}$ in 76 iterations. With TOI-Velocity it stays above $10^{-2}$, a miss of 17 to 21\,cm, through 3000 iterations while TOI-Velocity with spin stops at a miss of 2.8\,cm.}}
  \label{fig:motivating}
\end{figure}
We introduce an optimal control problem related to the game of pinball to highlight the gaps in existing approaches to differentiable contact resolution for non-spherical bodies. A thin elliptical flipper is pinned at its base, and a ball approaches it at an angle. The flipper rests at a fixed angle until a flip time $t_f$ and then swings at a constant angular velocity $\omega$. The goal is to choose the control $(t_f,\omega)$ so that the ball, after it is struck, reaches a target position at a final time $T$, and the loss is the squared distance between the ball's center and the target at time $T$. We formulate this as an optimal control problem in Sec.~\ref{sec:exp-control}. The target is generated from a known control, so the optimal loss is zero and the optimal control is known.
We solve the problem with vanilla gradient descent in a simulator with time step $\Delta t=10$\,ms, using either the gradients of TOI-Velocity \citep{zhong2023improving} or exact gradients using our approach.

Fig.~\ref{fig:motivating}(a) shows the setup and Fig.~\ref{fig:motivating}(b) shows the true loss during gradient descent (both from the same initial condition). With TOI-Velocity, the loss barely decreases, and after 3000 iterations the control still misses the target by 18\,cm. Changing the learning rate does not help: the iterates either oscillate between detection steps or diverge to controls where the ball misses the flipper. TOI-Velocity rewinds to the impact with a linear approximation that uses only the linear velocities and ignores the flipper's spin. Adding the spin to this approximation reduces the miss to 2.8\,cm but does not remove it (Fig.~\ref{fig:motivating}b). A one-step linear estimate of the impact time is inexact for an oblique collision with a curved body, which biases both the simulated outcome and its gradient (Sec.~3). With our exact gradients, gradient descent recovers the optimal control.

\subsection{Problem Setup}

We consider a fixed-step, impulse-based simulator with time step $\Delta t$. At each step, the simulator advances the bodies, checks whether they overlap at the end of the step, and, if they do, resolves the contact with an impulse computed from the contact point and normal. Gradients of a loss with respect to controls or parameters are obtained by differentiating through this sequence. The contact point, normal and impulse should be evaluated at the time of impact $t^\star$, the first time the bodies touch. Our goal is to compute the time of impact and its derivatives exactly for ellipsoids, so that the simulator's gradients are those of the true dynamics.

\section{Time-of-Impact (TOI) Rewinds and Why They Fail}\label{sec:background}

Algorithm~\ref{alg:step} shows one step of a typical fixed-step, impulse-based simulator. If the bodies overlap at the end of the step, the contact point and normal are computed from the overlapping state, denoted by \textsc{ContactNormal}, and an impulse updates the velocities. Here \textsc{ImpulseModel} maps the state, contact point and normal to a velocity impulse, for example the closed-form frictionless restitution impulse, or a convex program with friction \citep{zhong2021extending}; our method does not depend on this choice. Gradients are obtained by differentiating through these operations. Let $t_k$ be the first step end at which the bodies overlap, and let $t^\star\in(t_k-\Delta t,\,t_k]$ be the time of impact. Without a correction, the impulse is applied at $t_k$ using the geometry of the overlapping configuration. However, the true contact occurred at an earlier time $t^\star$ and by the end of the step, the objects have continued moving, changing their relative orientation leading to incorrect gradients.
\begin{algorithm}[b!]
  \caption{One step of a fixed-step, impulse-based simulator.}\label{alg:step}
  \KwIn{state $(x_k, v_k)$ at $t_k$, control $u$}
  $(x_{k+1}, v_{k+1}) \leftarrow \textsc{solve}(x_k, v_k, u, \Delta t)$\;
  \If{the bodies overlap at $t_{k+1}$}{
    $(p, n) \leftarrow$ \textsc{ContactNormal}($x_{k+1}$)\;
    $\Delta v \leftarrow \textsc{ImpulseModel}(x_{k+1}, v_{k+1}, p, n)$\;
    $v_{k+1} \leftarrow v_{k+1} + \Delta v$\;
  }
  \KwOut{$(x_{k+1}, v_{k+1})$}
\end{algorithm}

The existing approach \citep{hu2020difftaichi,zhong2023improving} to tackle this issue is based on a linear approximation of the distance between the two colliding bodies. Let $\phi(t)$ be a gap function between the bodies along their motion: $\phi>0$ when they are separated, $\phi=0$ when they touch, and $\phi<0$ when they overlap. The time of impact $t^\star$ is the first zero of $\phi$. For example, $\phi$ could be the relative signed distance function between the two rigid bodies. TOI-Velocity \citep{zhong2023improving} estimates $t^\star$ by a linear correction:
\begin{equation}\label{eq:rewind}
  \hat t = t_k - \frac{\phi(t_k)}{\phi'(t_k)}.
\end{equation}
This is the penetration depth divided by the closing speed, i.e., one Newton step on $\phi(t)=0$ from $t_k$. TOI-Velocity computes the closing speed from the linear velocities of the bodies, with their orientations held fixed. The centres of the bodies are moved back along their linear velocities to $\hat t$, their orientations are kept at their values at $t_k$, the impulse is applied there, and the bodies move forward with the new velocities for the rest of the step. A natural extension, which we call TOI-Velocity with spin, includes the angular velocities in the closing speed and rewinds the orientations of the bodies along with their centres. It is still one Newton step on $\phi(t)=0$ from $t_k$.

This estimate assumes the gap between bodies closes at a constant rate, which includes head-on collisions. When the collision is oblique and the bodies are curved at contact or when the bodies rotate, the gap closes at a varying rate, yielding incorrect impact times, contact points, normals, and the gradients computed through them. The error grows with the curvature of the bodies at the contact, the obliqueness of the collision, and the time step which we characterize below.

\begin{lemma}[Gradient error of the rewind]\label{lem:rewind}
Let two bodies first touch at $t^\star$, and let their gap $\phi(t;\theta)$ be $C^3$ in $(t,\theta)$ near the touch and close at a nonzero rate with $\phi'(t^\star)<0$. Suppose the overlap is detected at $t_k=t^\star+s$ with $0<s<\Delta t$. Then
\begin{equation}\label{eq:grad-error}
  \frac{d\hat t}{d\theta} = (1+\varepsilon)\,\frac{dt^\star}{d\theta} + O(\Delta t^2),
\end{equation}
where $\varepsilon = s\,\phi''(t^\star)/|\phi'(t^\star)|$ is the relative error in $dt^\star/d\theta$.
\end{lemma}
\begin{proof}[Proof]
The rewind $\hat t$ is a Newton step from $t_k$. We differentiate it with respect to $\theta$ and expand the result about $t^\star$ to prove the result. We use the notation $\phi_\theta=\partial\phi/\partial\theta$ and $\phi'_\theta$ similarly. Note that because $0<s<\Delta t$, small changes in $\theta$ do not change $t_k$. Differentiating gives
\[
  \frac{d\hat t}{d\theta} = -\frac{\phi_\theta}{\phi'} + \frac{\phi\,\phi'_\theta}{\phi'^2}\quad\text{at } t_k.
\]
At $t^\star$, $\phi=0$, and by the implicit function theorem, $dt^\star/d\theta = -\phi_\theta/\phi'$. Next, since $\phi$ is $C^3$, Taylor's theorem about $t^\star$ gives, with the right-hand sides evaluated at $t^\star$,
\begin{align*}
  \phi(t_k) &= s\,\phi' + O(s^2), & \phi'(t_k) &= \phi' + s\,\phi'' + O(s^2),\\
  \phi_\theta(t_k) &= \phi_\theta + s\,\phi'_\theta + O(s^2), & \phi'_\theta(t_k) &= \phi'_\theta + O(s).
\end{align*}

Finally, substituting these into the expression for $d\hat t/d\theta$, the $\phi'_\theta$ terms cancel and
\[
  \frac{d\hat t}{d\theta} = \Big(1-\frac{s\,\phi''}{\phi'}\Big)\Big(-\frac{\phi_\theta}{\phi'}\Big) + O(s^2) = (1+\varepsilon)\,\frac{dt^\star}{d\theta} + O(\Delta t^2),
\]
since $\phi'<0$ and $s<\Delta t$.
\end{proof}

\begin{corollary}\label{cor:rewind}
If the bodies are planar, strictly convex, translate with constant relative velocity $v$, and $\phi$ is their signed distance, then
\begin{equation}\label{eq:eps-geometry}
  \varepsilon = \frac{|v|\,s\,\sin\beta\,\tan\beta}{\rho_A+\rho_B},
\end{equation}
where $\rho_A$ and $\rho_B$ are their radii of curvature at the contact point and $\beta$ is the angle between $v$ and the contact normal ($\beta=0$ is head-on).
\end{corollary}
\begin{proof}[Proof sketch]
We assume without loss of generality that $A$ is stationary and $B$ translates with velocity $v$. We compare the bodies with their osculating discs at the contact, of radii $\rho_A$ and $\rho_B$. Since $\phi'(t^\star)$ depends only on the contact normal (only the velocity component along the normal affects the gap), and $\phi''(t^\star)$ only on how fast the normal turns, which is set by the curvatures, both values are the same as for the two discs. For two discs with centres $c_A$ and $c_B(t)$, the gap is $\phi(t)=|c_B(t)-c_A|-(\rho_A+\rho_B)$. Differentiating twice gives $\phi'=u\cdot v$ and $\phi''=(|v|^2-(u\cdot v)^2)/|c_B-c_A|$, where $u$ is the unit vector from $c_A$ to $c_B$. At $t^\star$, $|c_B-c_A|=\rho_A+\rho_B$ and $u$ is the contact normal $n$, so
\[
  \phi'=-|v|\cos\beta,\qquad \phi''=\frac{|v|^2\sin^2\beta}{\rho_A+\rho_B}.
\]
Substituting these into $\varepsilon=s\,\phi''/|\phi'|$ gives \eqref{eq:eps-geometry}.
\end{proof}

By Lemma~\ref{lem:rewind} and Corollary~\ref{cor:rewind}, the error $\varepsilon$ grows with the relative speed $|v|$, with the curvature of the bodies at the contact (as $\rho_A+\rho_B$ decreases), and with the obliqueness of the collision: it is zero for head-on collisions and grows without bound as $\beta$ approaches $90^\circ$. It also grows linearly with the time step, since $s\le\Delta t$. Our experiments in Sec.~\ref{sec:exp-gradients} show the same trends in the gradient error as the incidence angle, the aspect ratio of the body and the time step increase.

\section{Solution Approach}\label{sec:approach}

Lemma~\ref{lem:rewind} shows that the error of the rewind comes from estimating $t^\star$ from a single step. We instead compute $t^\star$ exactly and differentiate it exactly. Algorithm~\ref{alg:ours} shows one step of the resulting simulator. Continuous collision detection (CCD) finds the value of $t^\star$, and we do not differentiate it (Section~\ref{sec:ccd}). Every other operation in the step is a differentiable function: \textsc{solve} advances the bodies, \textsc{ToiGradient} attaches the derivative $dt^\star/d\theta$ to $t^\star$ (Section~\ref{sec:ift}), \textsc{ContactNormal} returns the contact point and normal (Section~\ref{sec:contact}), and \textsc{ImpulseModel} is the impulse model of Section~\ref{sec:background}. Automatic differentiation through these functions gives the derivatives of the step.

\begin{algorithm}[ht]
  \caption{One step with the exact time of impact.}\label{alg:ours}
  \KwIn{state $(x_k, v_k)$ at $t_k$, control $u$}
  $t^\star \leftarrow \textsc{CCD}(x_k, v_k, u)$ on $[t_k, t_{k+1}]$ \tcp*{not differentiated}
  \eIf{no contact was found}{
    $(x_{k+1}, v_{k+1}) \leftarrow \textsc{solve}(x_k, v_k, u, \Delta t)$\;
  }{
    $dt^\star/d\theta \leftarrow \textsc{ToiGradient}(t^\star, x_k, v_k, u)$ \tcp*{Section~\ref{sec:ift}}
    $(x^\star, v^\star) \leftarrow \textsc{solve}(x_k, v_k, u, t^\star - t_k)$ \tcp*{state at the time of impact}
    $(p, n) \leftarrow \textsc{ContactNormal}(x^\star)$ \tcp*{Section~\ref{sec:contact}}
    $\Delta v \leftarrow \textsc{ImpulseModel}(x^\star, v^\star, p, n)$\;
    $(x_{k+1}, v_{k+1}) \leftarrow \textsc{solve}(x^\star, v^\star + \Delta v, u, t_{k+1} - t^\star)$\;
  }
  \KwOut{$(x_{k+1}, v_{k+1})$}
\end{algorithm}

\subsection{Finding the time of impact}\label{sec:ccd}
Here we describe how to find the exact time of impact $t^\star$. Let $\alpha(t;\theta)$ be the minimum uniform scaling of the two bodies about their centers at which they intersect \citep{tracy2023differentiable}, where $\theta$ denotes the differentiation parameters, such as controls, initial states or shapes. The bodies are separated when $\alpha>1$, touch when $\alpha=1$, and overlap when $\alpha<1$; so we define the gap $\phi(t;\theta):=\alpha(t;\theta)-1$. Then $t^\star$ is the first zero of $\phi$ in the step. Any CCD algorithm that returns this zero to a given tolerance can be used since we do not differentiate through it and treat it like a black-box solver. We use conservative advancement: from $t_k$, we repeatedly advance time by a lower bound on the distance between the bodies divided by an upper bound on the relative speed of any two of their points. Since the distance cannot shrink faster than this speed, no step passes the first contact. Once the bodies are close, we refine $t^\star$ with Newton's method on $\phi(t)=0$.

\subsection{Differentiating the time of impact}\label{sec:ift}

We now describe \textsc{ToiGradient} in Algorithm~\ref{alg:ours}, which computes $dt^\star/d\theta$ at the value of $t^\star$ returned by CCD. The idea is to leverage the implicit function theorem and the Perram-Wertheim contact function \citep{perram1985statistical} for ellipsoids. The time of impact is defined implicitly by the gap $\phi(t^\star(\theta);\theta)=0$. If $\partial\phi/\partial t\neq0$ at $t^\star$, the implicit function theorem gives
\begin{equation}\label{eq:ift}
  \frac{dt^\star}{d\theta} = -\Big(\frac{\partial\phi}{\partial t}\Big)^{-1}\frac{\partial\phi}{\partial\theta}.
\end{equation}
Here $\partial\phi/\partial t$ is the rate at which the gap closes. It is nonzero unless the contact is grazing, and \eqref{eq:ift} grows without bound as the contact approaches grazing.

We briefly discuss how to compute these derivatives. Recall from Section~\ref{sec:ccd} that $\phi=\alpha-1$, where $\alpha$ is the minimum scaling at which the bodies intersect. Let $x(t;\theta)$ denote the configuration of the bodies at time $t$ so that $\phi(t;\theta)=\alpha(x(t;\theta);\theta)-1$, and let $x^\star$ be the state at the time of impact. By the chain rule,
\begin{equation}\label{eq:phi-partials}
  \frac{\partial\phi}{\partial\theta} = \frac{\partial\alpha}{\partial x}\frac{\partial x}{\partial\theta} + \frac{\partial\alpha}{\partial\theta},
  \qquad
  \frac{\partial\phi}{\partial t} = \frac{\partial\alpha}{\partial x}\frac{dx}{dt},
\end{equation}
with all derivatives evaluated at $t^\star$ and $x^\star$. The derivative $\frac{\partial x}{\partial\theta}$ is computed using automatic differentiation through $\textsc{solve}$ and $v^{\star} = \left.\frac{dx}{dt}\right|_{t=t^\star}$ is the velocity of the bodies just before impact.

Next we leverage the result of \cite{perram1985statistical} on the contact function for ellipsoids. Let body $i$ be defined by $\{y:(y-c_i)^\top S_i^{-1}(y-c_i)\le1\}$ with $S_i=R_i\,\mathrm{diag}(a_i)^2R_i^\top$, where $c_i$ is its centre, $R_i$ its rotation and $a_i$ its semi-axes. Then, for two ellipsoids $A$ and $B$,
\begin{equation}\label{eq:pw}
  \alpha^2 := \max_{\lambda\in[0,1]}F(\lambda),
  \qquad
  \text{with}\ F(\lambda)=\lambda(1-\lambda)\,r^\top G(\lambda)^{-1}r,
\end{equation}
where $r=c_B-c_A$ and $G(\lambda)=(1-\lambda)S_A+\lambda S_B$. Though $\alpha$ is the value of an optimization problem, we can leverage the envelope theorem to compute its partial derivatives. First, note that the function $F$ is strictly concave \citep{perram1985statistical} in $\lambda$ and zero at $\lambda=0$ and $\lambda=1$, so its maximizer $\lambda^\star$ is unique and lies in $(0,1)$. We find it by bisection on the sign of $\partial F/\partial\lambda$. Now, since $\partial F/\partial\lambda=0$ at $\lambda^\star$, the envelope theorem gives $\partial(\alpha^2)/\partial\xi =\partial F/\partial\xi$ at $\lambda^\star$ for any coordinate $\xi$ of $x$ or $\theta$. Since we have a closed-form expression for $F$, the derivative can be computed via auto-differentiation. In summary, $\alpha = \sqrt{F(\lambda^\star)}$ where the maximizer $\lambda^\star$ is computed via bisection on the derivative of $F$.

\subsection{Contact point and normal}\label{sec:contact}

The last piece we describe is \textsc{ContactNormal} in Algorithm~\ref{alg:ours}, which returns the contact point $p$ and the unit normal $n$ at the time of impact, together with their derivatives. Let $g=G(\lambda^\star)^{-1}r$. The contact point is given by \citep{perram1985statistical}:
\begin{equation}\label{eq:contact-point}
  p = c_A + (1-\lambda^\star)\,S_A\,g = c_B - \lambda^\star S_B\,g
\end{equation}
and the outward normal is $n=g/\|g\|$, pointing from $A$ to $B$.

The point and the normal depend on $\lambda^\star$ directly and we need the derivative of $\lambda^\star$. We obtain it the same way as for $t^\star$. We get $\lambda^\star$ using bisection, and its derivative follows from the implicit function theorem applied to the optimality condition $\partial F/\partial\lambda=0$:
\begin{equation}\label{eq:dlam}
  \frac{d\lambda^\star}{d\xi} = -\Big(\frac{\partial^2 F}{\partial\lambda^2}\Big)^{-1}\frac{\partial^2 F}{\partial\lambda\,\partial\xi}
\end{equation}
for any coordinate $\xi$ of $x$ or $\theta$. This is well defined because $F$ is strictly concave, so $\partial^2F/\partial\lambda^2<0$. With this, our solution approach is fully described and we proceed to discuss experimental results.

\section{Experiments}\label{sec:experiments}

We design the experiments to answer three questions. First, does the incorrect gradient of the time-of-impact rewind affect the ability to learn controllers? We study this on the pinball task of Sec.~\ref{subsec:motivation} over multiple targets (Sec.~\ref{sec:exp-control}). Second, how incorrect are the gradients, and how does the error depend on the obliqueness of the collision, the shape of the bodies and the time step? We measure this on a single impact against a ground truth (Sec.~\ref{sec:exp-gradients}). Third, can the exact gradients identify physical parameters from observed collisions? We estimate the shape and the coefficient of restitution of an ellipse from trajectory data (Sec.~\ref{sec:exp-sysid}). In all experiments, TOI-Velocity \citep{zhong2023improving} and its extension with spin are the baseline.

\subsection{Optimal Control Experiment}\label{sec:exp-control}
We consider a pinball flipper that must strike a ball so it reaches a target at a final time $T$. The flipper is an ellipse with semi-axes $a\ge b$ and moment of inertia $I_P$ about a pivot at one end of its major axis, and $\vartheta(t)$ denotes its angle. The ball is a disc of radius $R$ and mass $m_b$ with centre $c_b(t)$, initial position $c_0$ and initial velocity $v_0$. The control is $u=(t_f,\omega)$: the flipper rests at the angle $\vartheta_{\mathrm{rest}}$ until the flip time $t_f$ and then rotates with constant angular velocity $\omega$ (see Figure~\ref{fig:motivating}). The problem is formulated as an optimal control problem in continuous time with a state jump at the impact:
\begin{align}
\min_{t_f,\,\omega}\quad & \|c_b(T)-p_{\mathrm{target}}\|_2^2 \label{eq:ocp}\\
\text{subject to}\quad & \vartheta(t)=\vartheta_{\mathrm{rest}}+\omega\,\max(0,\,t-t_f), && t\le t^\star, \nonumber\\
& t^\star=\min\{t:\ \phi(t)=0\}, \nonumber\\
& c_b(t)=c_0+v_0\,t, && t\le t^\star, \nonumber\\
& c_b(t)=c_b(t^\star)+v_b^+\,(t-t^\star), && t>t^\star, \nonumber\\
& v_b^+=v_0-(1+e)\,\frac{v_n}{m_b\,k}\,n. \nonumber
\end{align}
Here $\phi$ is the gap between the flipper and the ball (Sec.~\ref{sec:ccd}): it is positive while they are apart and zero when they touch, so $t^\star$ is the impact time.
The ball moves in a straight line before and after the impact.
At $t^\star$, $p$ is the contact point (as in Algorithm~\ref{alg:step}), $n$ the unit contact normal, $\ell=p-q$ the lever arm from the pivot $q$, $v_n=n^\top(v_0-\omega\,\ell^\perp)$ the normal relative velocity, and $k=1/m_b+(\ell\times n)^2/I_P$. The flipper is body $A$ and the ball is body $B$, so $n$ points from the flipper to the ball.
The last constraint is the frictionless impulse for a flipper pinned at its pivot, with coefficient of restitution $e$.
The target is the position of the ball at time $T$ under a chosen control $u_{\mathrm{gen}}$, $p_{\mathrm{target}}=c_b(T;u_{\mathrm{gen}})$, so the optimal loss is zero and is attained at $u_{\mathrm{gen}}$.

To solve the above problem approximately, we discretize it in time with a step $\Delta t=T/N$:
\begin{align}
\min_{t_f,\,\omega}\quad & \|c_{b,N}-p_{\mathrm{target}}\|_2^2 \label{eq:ocp-discrete}\\
\text{subject to}\quad & s_{i+1}=\mathrm{step}(s_i,\,u,\,\Delta t),\qquad i=0,\dots,N-1, \nonumber
\end{align}
where $s_i=(\vartheta_i,c_{b,i},v_{b,i})$ is the state at time $t_i=i\,\Delta t$, and $\mathrm{step}$ is Algorithm~\ref{alg:step} with the impact handled either by the TOI-Velocity rewind \eqref{eq:rewind} or by Algorithm~\ref{alg:ours}.
We use $a=0.5$, $b=a/16$, $R=0.08$, $|v_0|=4$\,m/s, $e=1$, $T=1$\,s and $\Delta t=10$\,ms.

We run this experiment for 20 targets, each generated from a control $(t_f,\omega)$ drawn uniformly from $[0,0.2]\,\mathrm{s}\times[2,10]\,\mathrm{rad/s}$. Each run starts from a control that misses its target by about 20\,cm. For each target and method, we run gradient descent with five learning rates and report the best. Table~\ref{tab:targets} summarizes the results. Using our gradients, gradient descent reaches the optimal control for all 20 targets within a median of 144 iterations. Conversely, TOI-Velocity failed to reach any target within 3000 iterations, yielding a median final miss of 7.2 cm and worsening the control for 5 targets. Although TOI-Velocity with spin correction achieved a much smaller median miss of 1.2 mm (with 10 runs within 1 mm), none converged to the optimum, as gradient descent stopped where its gradient vanished (which is not the optimum). Results are summarized in Table~\ref{tab:targets}.
\begin{table}[t]
  \centering
  \caption{\small{Gradient descent on 20 targets, each method at its best learning rate per target. Within 1\,mm: final miss below 1\,mm. Converged: the final control equals the optimal control $u_{\mathrm{gen}}$ to within $10^{-10}$ in scaled units.}}
  \label{tab:targets}
  \small
  \begin{tabular}{lcccc}
    \hline
    Method & Within 1\,mm & Converged & Final miss (cm), & Iterations, \\
           & & & median [range] & median [range] \\
    \hline
    Ours & 20/20 & 20/20 & $4\times10^{-9}$ [$2\times10^{-9}$, $10^{-8}$] & 144 [71, 452] \\
    TOI-Velocity & 0/20 & 0/20 & 7.2 [0.8, 76] & 3000 [3000, 3000] \\
    TOI-Velocity with spin & 10/20 & 0/20 & 0.12 [0.003, 12] & 3000 [3000, 3000] \\
    \hline
  \end{tabular}
\end{table}

\subsection{Analysis of TOI-Velocity Gradients}
\label{sec:exp-gradients}

We measure how accurately TOI-Velocity computes gradients through a single impact as the collision becomes oblique, the body becomes non-spherical, and the time step grows.
The setup is as follows. A disc of radius $R=0.02$ moves at constant velocity with speed 2\,m/s and strikes a fixed ellipse with semi-axes $a=0.5$ and $b\le a$; the collision is frictionless and elastic.
The parameters $\theta$ are the disc's initial position and velocity, and the output is its position $p(T)$ at a time $T$ after the impact.
We compare the Jacobian $dp(T)/d\theta$ with a ground truth obtained by central finite differences, and report the relative error $\|J-J_{\mathrm{true}}\|_F/\|J_{\mathrm{true}}\|_F$.
We vary three quantities, one at a time, around a base case: the incidence angle $\beta$ between the disc's velocity and the contact normal, from nearly head-on ($\beta=5^\circ$) to near grazing ($\beta=85^\circ$); the aspect ratio $a/b$, from 1 (a circle) to 16; and the time step $\Delta t$, from 0.5 to 20\,ms.
The base case is $\beta=45^\circ$, $a/b=8$ and $\Delta t=10$\,ms.
We run this for 50 trials where the impact time is spread uniformly within one time step (TOI-velocity depends on this), and we report the median and the 10th--90th percentile range.

\begin{figure}[t]
  \centering
  \includegraphics[width=\linewidth]{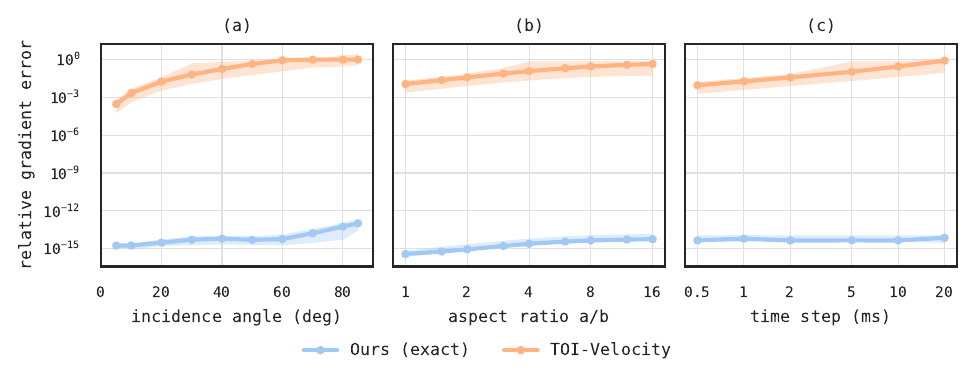}
  \caption{\small{Relative error of the gradient through a single impact against (a) the incidence angle, (b) the aspect ratio of the ellipse, and (c) the time step. Lines are medians and bands the 10th--90th percentiles over 50 impact times within a step.}}
  \label{fig:grad-accuracy}
\end{figure}

Fig.~\ref{fig:grad-accuracy} shows TOI-Velocity's gradient error as each quantity is varied around the base case.
The error is small for nearly head-on collisions (0.03\% at $5^\circ$) and grows with the incidence angle, to 18\% at $40^\circ$ and close to 100\% from $60^\circ$ on (Fig.~\ref{fig:grad-accuracy}a).
At $80^\circ$ and $85^\circ$, TOI-Velocity also misses the impact in 26 and 38 of the 50 cases: the disc overlaps the tip for less than one step, so no step end detects it.
The error grows with the aspect ratio, from 1.1\% for a circle to 45\% at $a/b=16$, as the contact point becomes more curved (Fig.~\ref{fig:grad-accuracy}b); a circle is not exact either, because the collision is oblique.
The error grows roughly in proportion to the time step, from 0.9\% at 0.5\,ms to 83\% at 20\,ms (Fig.~\ref{fig:grad-accuracy}c), so halving the step only halves the error.
For a fixed setting, the error varies by more than an order of magnitude depending on where in the step the impact falls.
Our gradients agree with the ground truth to within $3\times10^{-13}$ in every setting.

\subsection{System Identification}\label{sec:exp-sysid}

We estimate the shape and the coefficient of restitution of a body from observed collisions, as in DiffCoSim \citep{zhong2021extending}, where contact and system properties are learned from trajectories.
The setup is as follows. A free ellipse with semi-axes $a$ and $b=0.125$, unit mass and uniform density starts at rest at the origin, with a random orientation and an angular velocity drawn from $[-3,3]$\,rad/s. A ball of radius $0.1$ and mass $0.5$ approaches it at 2\,m/s from a random height and direction. The collision is frictionless with coefficient of restitution $e$. The unknown parameters are $\theta=(a,e)$.
The data are the positions of both bodies at $T=1.4$\,s for 12 such collisions, simulated with the exact time of impact at the true parameters. Each method estimates $\theta$ by minimizing the squared error between its simulated positions and the data, using Gauss--Newton with a backtracking line search, its own simulator and its own gradients.
We repeat this for 20 datasets, with true parameters drawn uniformly from $a\in[0.7,1.3]$ and $e\in[0.5,1]$, and start each fit at $\theta_{\mathrm{true}}-(0.15,0.15)$. We compare against TOI-Velocity and TOI-Velocity with spin at time steps from 2 to 20\,ms, and report the relative error of the estimates.

\begin{table}[t]
  \centering
  \caption{\small{System identification on 20 datasets: relative error (\%) of the estimated semi-axis $a$ and coefficient of restitution $e$, median [max] over the datasets. Ours does not depend on $\Delta t$.}}
  \label{tab:sysid}
  \small
  \begin{tabular}{llcccc}
    \hline
    Method & & $\Delta t=2$\,ms & 5\,ms & 10\,ms & 20\,ms \\
    \hline
    Ours & $a$, $e$ & \multicolumn{4}{c}{$<10^{-13}$} \\
    \hline
    TOI-Velocity & $a$ & 0.26 [15] & 0.45 [26] & 1.2 [17] & 2.7 [26] \\
                 & $e$ & 0.55 [30] & 1.1 [28] & 3.0 [25] & 4.5 [49] \\
    \hline
    TOI-Velocity with spin & $a$ & 0.003 [0.05] & 0.011 [0.19] & 0.088 [0.74] & 0.16 [4.0] \\
                           & $e$ & 0.003 [0.06] & 0.007 [0.24] & 0.084 [0.87] & 0.25 [4.3] \\
    \hline
  \end{tabular}
\end{table}

Table~\ref{tab:sysid} summarizes the results. First, our estimates are exact across all 20 datasets, recovering $a$ and $e$ within a relative error of $10^{-15}$. Second, TOI-Velocity's estimates are biased; the median error in $e$ grows from 0.55\% at 2\,ms to 4.5\% at 20\,ms (reaching 49\% on some datasets). On 12 to 18 of the 20 datasets, depending on $\Delta t$, the fitting stalls and no step along TOI-Velocity's gradient direction lowers its own loss. Third, TOI-Velocity with spin reduces but does not eliminate this error. Its median error in $e$ grows from 0.003\% at 2\,ms to 0.25\% at 20\,ms (up to 4.3\% peak) and its fit stalls on 3 to 7 of the 20 datasets.

\section{Related Work}\label{sec:related}

{\textbf{Differentiable contact models.}} Differentiable simulators handle rigid contact in several ways via velocity impulses from a complementarity problem \citep{deavila2018end,werling2021fast} or convex programs \citep{zhong2021extending}, compliant/position-based models \citep{freeman2021brax,macklin2022warp}, and interior-point smoothing \citep{howell2022dojo}. However, the gradients through contact can be inaccurate \citep{zhong2022differentiable}, and stiffness or discontinuities can make them less reliable than zeroth-order estimates \citep{suh2022differentiable}. Policy-learning methods often truncate or adapt the gradient horizon to avoid contact gradients \citep{xu2022accelerated,georgiev2024adaptive}, and better contact models for accurate gradients remain an open direction \citep{newbury2024review}. We focus on velocity-impulse simulators and impulse timing.

{\textbf{Time-of-impact corrections.}} DiffTaichi \citep{hu2020difftaichi} showed that applying the impulse at the end of the step gives wrong gradients for a ball hitting the ground and proposed an impact time estimate by dividing the penetration by the closing speed. TOI-Velocity \citep{zhong2023improving} used this estimated time to evaluate the velocity and the contact normal for collisions between balls. Both corrections are one Newton step on the gap from the detecting step end. \citet{yang2024jade} instead finds the impact time between meshes with continuous collision detection and splits the step there. It differentiates the impact time through an expression at the detected contact point. \citet{qiao2020scalable} resolve collisions between meshes in localized impact zones and differentiate the resulting optimization implicitly. We show that the one-step estimate is exact only when the gap closes at a constant rate, quantify its error, and replace it with the exact impact time for ellipsoids.

{\textbf{Event times in hybrid systems.}} In hybrid systems, the derivative of an event time follows from the implicit function theorem on the guard. Trajectory sensitivity analysis uses it to carry sensitivities across events and state jumps \citep{hiskens2000trajectory,saccon2014sensitivity}, and it enters the saltation matrix, the sensitivity update at a jump \citep{kong2024saltation}. Implicit differentiation of event times has also been used for neural ODEs with events \citep{chen2021learning}. For differential equations with discontinuous right-hand sides, the gradients of a time-stepping discretization need not converge to the true gradients as the step shrinks, even when the trajectories do \citep{stewart2010optimal}. In optimal control of rigid bodies with impacts and friction, FESD-J \citep{nurkanovic2024fesdj} lets the step lengths vary so that impacts fall on step boundaries. A fixed-step simulator cannot move its step boundaries, so we correct only the step that contains the impact.

\textbf{Collision detection and differentiation.} Continuous collision detection for ellipses and ellipsoids is well studied \citep{choi2006continuous,choi2009continuous}, and the Perram--Wertheim contact function gives an overlap measure for ellipsoids that requires only a one-dimensional maximization \citep{perram1985statistical}. \citet{donev2005neighbor} use this function in event-driven molecular dynamics to find exact collision times of moving ellipsoids, but do not differentiate them. For general convex bodies, the minimum uniform scaling at which two bodies touch is the growth distance \citep{ong1996growth}. DCOL \citep{tracy2023differentiable} makes collision detection differentiable by solving for this scaling between convex primitives, but it works at a single configuration and does not provide the impact time. We combine exact collision detection with implicit differentiation at the touch to obtain exact gradients through impacts between ellipsoids.

\section{Conclusion}\label{sec:conclusion}
We showed that the time-of-impact correction used in differentiable simulators for rigid body contact gives incorrect gradients in a variety of settings including non-spherical objects and oblique collisions. In a 2D pinball task, these errors prevent gradient descent from finding the optimal control. Replacing the estimate with the exact impact time from continuous collision detection, and its derivatives from the implicit function theorem, gives exact gradients and with them, gradient descent reached the optimal control. In a system identification task, the exact gradients recover the shape and the coefficient of restitution of an ellipse at every time step, while the estimates obtained with the time-of-impact corrections are biased. Our current approach is suited for ellipsoids but we believe the ideas can be applied to general convex bodies which is an avenue for future work.


\bibliographystyle{assets/plainnat}
\bibliography{paper}

\end{document}